# Multiple Object Tracking with Motion and Appearance Cues

Weiqiang Li, Jiatong Mu, Guizhong Liu
School of Information and Communication Engineering, Xi'an Jiaotong University, Xi'an, China
`lwq1230@stu.xjtu.edu.cn, m625329163@stu.xjtu.edu.cn, liugz@xjtu.edu.cn`

## Abstract

*Due to better video quality and higher frame rate, the performance of multiple object tracking issues has been greatly improved in recent years. However, in real application scenarios, camera motion and noisy per frame detection results degrade the performance of trackers significantly. High-speed and high-quality multiple object trackers are still in urgent demand. In this paper, we propose a new multiple object tracker following the popular tracking-by-detection scheme. We tackle the camera motion problem with an optical flow network and utilize an auxiliary tracker to deal with the missing detection problem. Besides, we use both the appearance and motion information to improve the matching quality. The experimental results on the VisDrone-MOT dataset show that our approach can improve the performance of multiple object tracking significantly while achieving a high efficiency.*

## 1. Introduction

Computer vision is an important branch of artificial intelligence, and multiple object tracking (MOT) has become a research hotspot in the field of computer vision. According to the review literature written by Luo *et al.* [1], the task of MOT is mainly partitioned to locating multiple objects, maintaining their identities, and yielding their individual trajectories given an input video. Compared with single object tracking (SOT), MOT pays more attention to the determination of the individual trajectories of multiple objects and it is a more complex issue due to interactions among multiple objects. According to Micheloni *et al.* [2], MOT has very important practical value in the fields of video surveillance, automatic driving, robot navigation and positioning, intelligent human-computer interaction, etc.

In recent years, with the rapid development of deep neural network, the accuracy of object detection has risen to a new level. As a result, tracking-by-detection has become the most popular framework for multiple object tracking (MOT). First of all, a detector is used to detect all the objects in each frame. Then the data association method is used to obtain the respective trajectory of each object.

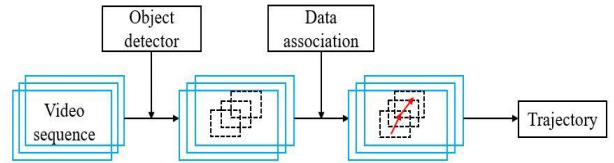

Figure 1. The main procedure of tracking-by-detection framework.

Figure 1 shows the main procedure of tracking-by-detection. Under this process, the performance of MOT depends largely on the quality of the object detection algorithm.

Simple tracking-by-detection method like IoU Tracker proposed by Bochinski *et al.* [3] can achieve a fine result when the object detection results are good enough and there is no dramatic camera motion in the video. However, camera motion is very common and dramatic for videos taken by drones, and the processing of videos taken by drones is also in urgent demand. Not only that, when the objects are crowded and there are a lot of interactions in the scene, most object detectors are often difficult to deal with it and many false-positive detections and missing detections will occur at this time. These problems make multiple object tracking a more complicated challenge.

So far, many methods have been proposed to response to these problems. Wojke *et al.* [4] propose a deep association metric considering both the motion information and the deep appearance feature of the object while matching. Chen *et al.* [5] further improve the appearance feature and handle unreliable detection by collecting candidates from outputs of both detection and tracking. Chu *et al.* [6] apply single object tracking method to multiple object tracking issues and propose a spatial-temporal attention mechanism to handle the drift caused by occlusion and interaction among targets. Tang *et al.* [7] propose a novel graph-based formulation that links and clusters person hypotheses over time by solving an instance of a minimum cost lifted multicut problem.

In this paper, we mainly solve the problems of ID switches and error detections in multiple object tracking from three aspects and propose a new method of MOT named Flow-Tracker. For frequent camera motion in videos taken by drones, we use the optical flow network proposed by Sun *et al.* [8] to eliminate its influence and estimate the global motion of two adjacent frames. It also acts as a tracker to predict the position of the object in the current frame, which is more favorable for the subsequent data association process. Second, we propose a cascade matching strategy based on IoU and deep appearance features, which has a good effect on reducing false matches. In addition, we utilize the optical flow network as an auxiliary tracker when the trajectory is broken due to the missing detection. It has a great effect on alleviating the problems of ID switches and fragmentations caused by missing detection. The experiments on the VisDrone2019-MOT dataset [9] show that our method can improve the accuracy of multiple object tracking significantly. Further, we can achieve a high speed of 100 FPS with performing motion estimation by judging that each frame occurs camera motion or not, which can achieve a trade-off between the accuracy and the speed.

## 2. Related work

The research of multiple object tracking (MOT) problem has been a long time. In recent years, the problems of object detection and tracking under the UAV scenes has aroused the attention of researchers. More and more large-scale datasets based on drones are also appearing, such as Stanford Drone Dataset (SDD) [10], DTB70 dataset [11], VisDrone dataset [9] and so on. In order to tackle the various challenges of MOT under drone scenes, we need to consider the effective use of the motion and appearance information, better data association strategy and more accurate object detectors, etc. Many related works have thoroughly studied about these issues.

### 2.1. Motion estimation

The task of object tracking is to predict the position of the object. Due to the dramatic camera motion under the drone scenes, the prediction becomes more complicated. In some earlier works, the Kalman filter [12] is a commonly used motion estimation method in MOT, predicting the target state of the current moment from the target state at the previous moment. Recently, with the development of deep learning, the motion models [13, 14] based on RNN and the Long Short Term Memory (LSTM) have achieved better results.

The optical flow is an effective way to describe motion between frames within a video. The traditional Lucas–Kanade algorithm [15] gives a method for solving sparse optical flow, which has been widely used. With the explosive progress of convolutional neural network, the method of estimating the optical flow directly by CNN has also been proposed. Fischer and Ilg *et al.* successively propose FlowNet [16] and FlowNet 2.0 [17], which can predict the optical flow directly using a well-trained encoder-decoder network and can be used for dense optical flow estimation. Sun *et al.* propose PWC-Net [8], an optical flow network fusing pyramidal processing, warping, and a cost volume, which has achieved better and faster optical flow estimation. Our algorithm takes it as the way of motion estimation in the process of MOT.

### 2.2. Appearance feature

The appearance feature is a more discriminative representation of the object, which can distinguish between objects effectively when they are similar. It is very helpful for crowded objects and scenes where there are lots of interactions among objects. In earlier works, the color histograms [18, 19] and some hand-crafted features [20, 21] are commonly used as descriptors of the appearance of objects.

With the popularity of deep neural network, deep feature based appearance representations are increasingly used to enhance the discriminative power of appearance features. Wojke *et al.* [4] employ a wide residual network to extract the features of objects and measure the similarity of objects with cosine distance. Chen *et al.* [5] utilize the network architecture proposed by Zhao *et al.* [22] and train the network on a combination of several large-scale person re-identification datasets to extract the features of objects, which takes Euclidean distance as the metric of similarity of objects. Leal-Taixé *et al.* [23] extensively use Siamese network to learn discriminative features from detected objects. In this paper, we extract the appearance features of the detected objects using a residual network trained on large-scale re-identification datasets and distinguish them by calculating the cosine distance between two objects.

### 2.3. Data association

Data association is a key step in tracking-by-detection based MOT methods. Many offline MOT methods [24, 25, 26] treat data associations as graph-based optimization problems. Hungarian algorithm [27] is another commonly used data association optimization method. Xu *et al.* [28] further introduce a differentiable operator to build a deep Hungarian network.

We simply replace the greedy data association way in IoU Tracker [3] with the Hungarian algorithm. In addition, we design a cascade data matching method by repeatedly utilizing the motion information and appearance features of the objects.

### 2.4. Object detection

As a part of tracking-by-detection based MOT algorithm, object detection has a great impact on the performance of the trackers. Both false positives and missing detections directly affect the evaluation metric of MOT, and indirectly lead to ID switches, so a better detector can greatly improve the accuracy of MOT. In earlier times, pedestrian or vehicle detectors based on DPM [29] played an important role in MOT. Recently, deep learning based object detection methods have far surpassed those traditional ones. Faster R-

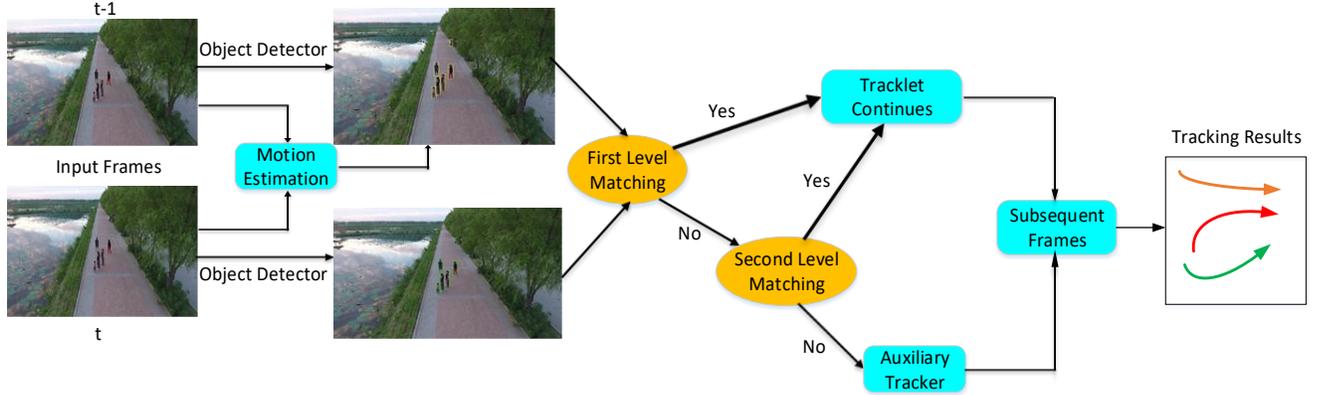

Figure 2. The overview of the proposed Flow-Tracker, which mainly contains three modules proposed in this paper. We employ an optical flow network for motion estimation to eliminate camera motion. A cascade matching policy is introduced to make full use of the motion and appearance information of the objects. And an auxiliary tracker is used to reduce mismatching caused by missing detections.

CNN [30] has become a commonly used object detector which can make good performance. Some recent object detection algorithms [31, 32, 33, 34, 35] continuously refresh the accuracy of object detection. Furthermore, some methods of pedestrian detection [36, 37] are also usually used as benchmark detectors for MOT.

Since the objects under drone shooting are small and crowded, we need realize better object detectors to improve the poor performance of MOT. We compare the tracking results of Faster R-CNN and several improved algorithms in this paper, showing the big impact of object detector on MOT.

## 3. Method

As mentioned above, camera motion and noisy detection results are main problems to be solved of high quality multiple object tracker, and our Flow-Tracker is designed to deal with these two challenges. It uses IoU Tracker as the baseline tracker and handles global motion problems caused by camera with an optical flow network, which reduces the amount of ID switches obviously. Against mismatching caused by missing detections, an auxiliary tracker and a better cascade matching strategy can effectively deal with it. Besides, we utilize more accurate detector to eliminate the effects of false alarms and missing detections. Figure 2 gives the overall framework and procedure of our proposed Flow-Tracker.

### 3.1. IoU Tracker

We use IoU Tracker as the baseline due to its simplicity and high efficiency. The IoU Tracker takes advantages of the high quality and high frame rate of videos. It only uses IoU as the matching criteria of objects in two adjacent frames, which is defined as:

$$\text{IoU}(bbox1, bbox2) = \frac{bbox1 \cap bbox2}{bbox1 \cup bbox2} \quad (1)$$

IoU Tracker simply continues a track by associating the detection with the highest IoU to the tracked object in the previous frame if a threshold $\sigma_{IoU}$ is met, which is a greedy way. All detections not assigned will be created as new tracks. If a track does not have any detections to assign, it will be finished. In order to reduce the impact of false-positive detections, all finished tracks with a length shorter than $t_{min}$ and without at least one detection score above $\sigma_h$ are filtered. Figure 3 shows the main principle of IoU Tracker.

The whole tracking process is lightweight and efficient. When there is no camera motion in video sequences, IoU Tracker is a good multiple object tracker. However, camera motion will cause lots of errors in IoU-based matching method, further leading to ID switches. In addition, missing detections and false-positive detections are also two factors affecting the accuracy of association.

### 3.2. Global motion estimation

With the widespread use of drones, more and more videos are under the drone scenes. Therefore, camera motion has become a big challenge to MOT. When there is a large amount of camera motion in video sequences, large offsets will occur in the objects of two adjacent frames, which affects the accuracy of matching results.

In order to eliminate the effects of the camera motion, we need to compensate for the motion of two adjacent frames. We use the optical flow network (PWC-Net) proposed by Sun *et al.* [8] to estimate the amount of motion at each position from the previous frame $f_{t-1}$ to the current frame $f_t$. For each track in the previous frame $f_{t-1}$, we use the estimated offset from PWC-Net to calculate its exact position in the current frame $f_t$:

$$bbox'(x_1, y_1) = bbox(x_1 + u_1, y_1 + v_1) \quad (2)$$
$$bbox'(x_2, y_2) = bbox(x_2 + u_2, y_2 + v_2) \quad (3)$$

where $bbox$ and $bbox'$ are the bounding boxes of the same object in the previous frame and the current frame, respectively. $(x_1, y_1)$ and $(x_2, y_2)$ are the coordinates of the top left and bottom right corners of the object. $(u_1, v_1)$ and $(u_2, v_2)$ are respectively the horizontal and vertical optical flow values at the top left and bottom right corners of the object.

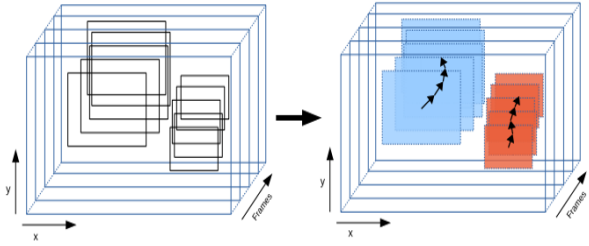

Figure 3. The main process of IoU Tracker. It takes IoU as the criterion for matching objects of adjacent frames, simple and efficient.

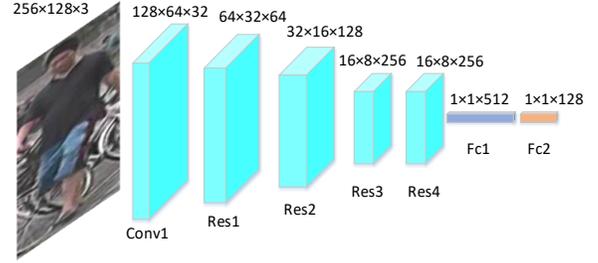

Figure 4. Overview of the deep appearance feature network, which is based on ResNet [39]. The feature of each detected object is repre-sented by a 128-dimensional vector.

In fact, the optical flow network estimates the amount of global motion between two adjacent frames, taking into account the effects of camera motion. We can also consider it as a predictor of the object position, predicting the object position in the current frame from the global motion amount estimated by the optical flow given the object position in the previous frame. As a result, we associate predicted objects by optical flow and the detected objects in the current frame, which is a more precise way.

Because camera motion does not occur in each frame of the whole video, we propose another method of motion estimation. We count the number of unmatched objects in the current frame, and if it exceeds half of the matched objects, we think this is caused by camera motion, so we need use optical flow estimation to predict the positions of objects at this time. The experiments show that it is a more efficient method which can reach a high speed of 100 FPS.

### 3.3. Auxiliary tracker

Another drawback of IoU Tracker is that the previous track cannot continue when there is missing detection in a certain frame. In this case, it will create a new track in the subsequent frames, which causes a large number of ID switches and fragmentations.

When the object disappears due to missing detection in a frame $f_t$, it may reappear in subsequent frames, so we cannot simply terminate this track. Instead, we utilize an auxiliary tracker which is actually a position predictor to predict the position of the object in subsequent several frames.

Specifically, we use the optical flow network mentioned in the previous section to predict the location of unmatched objects, which also saves lots of computational overhead. In order to prevent errors of prediction in more frames, we only limit the use of auxiliary tracker to a maximum of $t_{max}$ frames. Within these $t_{max}$ frames, the previous track is continued with the object bounding box predicted by the auxiliary tracker. If the track can be successfully matched with a new detection within these $t_{max}$ frames, it is considered that a missing detection has occurred before and the track will continue. Otherwise, we believe that the object has disappeared and this track will be terminated.

The auxiliary tracker is very helpful for reducing missing detections and fragmentations, which improves the matching quality effectively. And the experimental results confirm its effectiveness.

### 3.4. Cascade matching policy

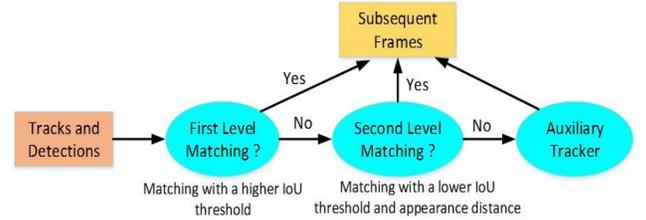

Figure 5. The procedure of cascade matching policy.

IoU Tracker only takes the IoU distance as the criterion for associating objects between adjacent frames. This may be inaccurate when there are crowded objects and a lot of false-positive detections in the scene, so we think we should utilize both IoU and the appearance feature to improve the accuracy of data association. We extract the appearance features of the detected objects using a residual network trained on large-scale re-identification datasets and distinguish two objects by calculating the cosine distance between them. Based on these, we further propose a cascade matching policy. Figure 4 and Figure 5 demonstrate our appearance feature extractor and the procedure of cascade matching policy respectively.

The specific matching process can be divided into three steps. First of all, we use optical flow network to predict the motion between two adjacent frames and derive the object position on the current frame. Then we calculate the IoU between the tracked object and the detected object. If it is above the defined threshold $\sigma_{IoU1}$, we think they are matched. Second, we extract the appearance features of unmatched tracks and detections. Then we compute the appearance distance and IoU between unmatched tracks and detections. If they meet the matching criteria at the same time, we think they are matched. Finally, for those mismatched tracks, we use the auxiliary tracker to continue predicting their positions in subsequent several frames. If they match successfully within these frames, we believe these tracks can continue. Otherwise, we think these objects have disappeared and these tracks will be terminated. We set a higher threshold of IoU in the first step and introduce the appearance features of the objects in the second step, combining together for a more accurate matching. Besides, the use of auxiliary trackers can reduce the impact of missing detections.

| Method | AP | $AP_{0.25}$ | $AP_{0.5}$ | $AP_{0.75}$ | $AP_{car}$ | $AP_{bus}$ | $AP_{trk}$ | $AP_{ped}$ | $AP_{van}$ |
|---|---|---|---|---|---|---|---|---|---|
| Deep Sort [42] | 4.27 | 7.14 | 4.05 | 1.62 | 12.17 | 0 | 1.04 | 7.22 | 0.9 |
| IoU Tracker [3] | 10.18 | 15.04 | 9.34 | 6.17 | **34.05** | 0 | 0 | **7.69** | 9.17 |
| Flow-Tracker(ours) | **15.12** | **26.03** | **11.60** | **6.25** | 32.82 | **6.08** | **9.72** | 7.65 | **9.84** |

Table 1. The performance of multiple object tracking on VisDrone2019-MOT validation set based on AP metric.

| Method | MOTA | MOTP | $IDF_1$ | MT | ML | FP | FN | IDS | FM |
|---|---|---|---|---|---|---|---|---|---|
| Deep Sort [42] | 10.1 | 74.7 | 38.3 | 106 | **245** | 21172 | 42826 | 590 | 1101 |
| IoU Tracker [3] | 12.6 | 75.7 | 38.3 | 113 | 248 | 19979 | **42236** | 576 | 1093 |
| Flow-Tracker(ours) | **26.4** | **78.1** | **41.9** | **115** | 246 | **9987** | 43766 | **127** | **428** |

Table 2. The performance of multiple object tracking on VisDrone2019-MOT validation set based on MOTA metric.

## 4. Experiments

We perform a lot of experiments on the VisDrone2019-MOT dataset and evaluate the performance of the proposed Flow-Tracker. We mainly compare with the baseline method using two different evaluation metrics and the experimental results confirm the effectiveness of our method.

### 4.1. Experiment setup

**VisDrone datasets.** VisDrone [9] is a large-scale benchmark under drone scenes, which contains four tasks of DET, VID, SOT and MOT. VisDrone2019-MOT dataset consists of 63 videos captured by drone platforms in different places, annotating the bounding boxes of ten categories (i.e., pedestrian, person, car, van, bus, truck, motor, bicycle, awning-tricycle, and tricycle) of objects in each video frame. The training set, validation set and test set contain 56, 7 and 16 videos respectively. All experiments in this paper are trained on the training set and we report the results on the validation set and the test set.

**Implementation details.** We employ PWC-Net trained on FlyingChairs [16] and FlyingThings3D [38] datasets as our motion estimation network and take it as the auxiliary tracker simultaneously when there is missing detection. Our appearance feature extractor is based on ResNet [39], which is pretrained on a combination of Market1501 [40] and MARS [41] datasets. We use three object detection algorithms: Faster R-CNN, Cascade R-CNN and improved Cascade R-CNN. For the object detector, we use a GTX1080Ti GPU to train it on all the images in the training set. And all the hyper-parameters in the experiments are obtained by grid search on the validation set.

**Evaluation metrics.** To evaluate the performance of different methods on MOT task, we adopt two evaluation ways.

1). Each algorithm outputs a list of bounding boxes with confidence scores and the corresponding identities. We sort the tracklets (formed by the bounding box detections with the same identity) according to the average confidence of their bounding box detections. A tracklet is considered correct if the IoU with ground truth tracklet is larger than a threshold. We use three thresholds in evaluation, i.e., 0.25, 0.50, and 0.75. The performance of an algorithm is evaluated by averaging the mean average precision (mAP) across object classes over different thresholds.

2). We also adopt the most commonly used metrics in MOT, including multiple object tracking accuracy (MOTA), multiple object tracking precision (MOTP), identification F1 score (IDF1), the number of mostly tracked targets (MT, > 80% recovered), the number of mostly lost targets (ML, < 20% recovered), false positives (FP), false negatives (FN) and identity switches (IDS). Besides, we also consider the processing speed of the algorithm and use frames per second (FPS) to measure it.

### 4.2. Results and analysis on validation set

We first use Faster R-CNN as the object detector and compare our method with Deep Sort [4] and IoU Tracker [3]. The results are shown in Table 1 and Table 2.

Specifically, we first perform a class-agnostic non-maximum suppression (NMS) with a threshold $\sigma_{nms}$ for all the detection results of each image. Then we employ the proposed improvements to our tracker. From the results of Table 1, we find the mean average precision (mAP) has a significant improvement than the baseline method and the accuracy of most categories has been improved, which proves the effectiveness of our method. Further, from Table 2, the MOTA of our method has a substantial increase compared to Deep Sort and IoU Tracker. Not only that, we can find the number of false positives has been greatly reduced. The number of ID switches and fragments are also greatly reduced, confirming that our proposed motion estimation module, auxiliary tracker and cascade matching strategy have improved the accuracy of matching.

| Method | AP | $AP_{0.25}$ | $AP_{0.5}$ | $AP_{0.75}$ | $AP_{car}$ | $AP_{bus}$ | $AP_{trk}$ | $AP_{ped}$ | $AP_{van}$ |
|---|---|---|---|---|---|---|---|---|---|
| IoU Tracker [3] | 10.18 | 15.04 | 9.34 | 6.17 | 34.05 | 0 | 0 | 7.69 | 9.17 |
| + Cascade R-CNN | 16.68 | 29.38 | 13.72 | 6.93 | 31.84 | 11.11 | 21.88 | 7.79 | 10.79 |
| + motion estimation | 17.59 | **30.50** | 14.32 | 7.26 | 33.95 | 11.11 | 21.88 | 8.99 | 12.04 |
| + auxiliary tracker | 19.46 | 28.72 | 19.10 | 12.87 | 42.14 | 11.11 | 21.81 | 11.16 | 19.75 |
| + cascade matching | 20.58 | 29.83 | 19.21 | 13.70 | 44.73 | 11.11 | 25.00 | 12.89 | 20.29 |
| Flow-Tracker | **21.70** | 30.30 | **20.09** | **15.72** | **46.78** | 11.11 | **25.00** | **13.94** | **22.69** |

Table 3. Comparison of multiple object tracking results on VisDrone2019-MOT validation set based on AP metric. From top to bottom, each row indicates the result of adding different modules proposed in this paper to the baseline tracker.

| Method | MOTA | MOTP | $IDF_1$ | MT | ML | FP | FN | IDS | FM |
|---|---|---|---|---|---|---|---|---|---|
| IoU Tracker [3] | 12.6 | 75.7 | 38.3 | 113 | 248 | 19979 | 42236 | 576 | 1093 |
| + Cascade R-CNN | 26.7 | 78.3 | 41.8 | 117 | 248 | 10179 | 42151 | 338 | 630 |
| + motion estimation | 29.0 | 78.3 | 42.8 | 121 | 246 | 9316 | 41608 | 290 | 574 |
| + auxiliary tracker | 31.2 | 78.6 | 45.7 | 136 | 253 | **9123** | 40334 | 221 | 542 |
| + cascade matching | 31.5 | 78.5 | 46.0 | 137 | 247 | 9547 | 39474 | 125 | 489 |
| Flow-Tracker | **32.1** | **78.7** | **50.1** | **141** | 240 | 9242 | **39423** | **112** | **475** |

Table 4. Comparison of multiple object tracking results on VisDrone2019-MOT validation set based on MOTA metric. From top to bottom, each row indicates the result of adding different modules proposed in this paper to the baseline tracker.

| Method | AP | MOTA | Speed (FPS) |
|---|---|---|---|
| Flow-Tracker | 21.7 | 32.1 | 5 |
| Flow-Tracker-fast | 20.9 | 31.6 | 100 |

Table 5. Comparison of accuracy and speed of the proposed two methods on the validation set of VisDrone2019-MOT. Flow-Tracker-fast is a way that we do not estimate the optical flow per frame.

Because the detection results of Faster R-CNN on the VisDrone dataset are not very good and there are still a lot of false positives and missing detections, which have influenced the correct association of objects. We use improved detection methods to improve the performance of tracker. The experimental results are presented in Table 3 and Table 4, and we analyze the role of different modules proposed in this paper.

**Effect of motion estimation.** We add a motion estimation module to predict the position of the object in the current frame before the object matching process. The results in Table 3 show that the overall AP has some improvement after adding it. In Table 4, we can see that the amount of false positives, missing detections and ID switches reduce significantly with our motion estimation module, which confirms that eliminating camera motion by using optical flow information has great help in reducing false matches.

**Effect of auxiliary tracker.** The overall AP can be raised from 17.59 to 19.46 by adding an auxiliary. Further, the false positives and missing detections reduce greatly thanks to the auxiliary tracker. Besides, the introduction of the auxiliary tracker significantly reduces the number of ID switches caused by missing detections, which also makes the fragmentations in a complete trajectory less. In general, it raises the MOTA by 2.2 points.

**Effect of cascade matching policy.** Our matching strategy not only considers IoU between objects, but also introduces appearance features to enhance the discrimination of the objects. The overall AP has already risen to 20.58 by using cascade matching strategy and the accuracy of each category has increased more or less. For another evaluation metric, the MOTA has a minor improvement which also states the effectiveness of our matching method. We can also find that the number of ID switches is reduced by up to 45% from Table 4. At the same time, the number of fragments in the trajectory is also significantly reduced, proving the importance of data association and matching in multiple object tracking.

**Effect of object detector.** We first use Faster R-CNN as the object detector of the original IoU Tracker. Then we train a Cascade R-CNN detector on the VisDrone2019-MOT training set and replace the original object detector.

| Method | AP | $AP_{0.25}$ | $AP_{0.5}$ | $AP_{0.75}$ | $AP_{car}$ | $AP_{bus}$ | $AP_{trk}$ | $AP_{ped}$ | $AP_{van}$ |
|---|---|---|---|---|---|---|---|---|---|
| CEM [42] | 5.7 | 9.22 | 4.89 | 2.99 | 6.51 | 10.58 | 8.33 | 0.7 | 2.38 |
| H$^2$T [43] | 4.93 | 8.93 | 4.73 | 1.12 | 12.9 | 5.99 | 2.27 | 2.18 | 1.29 |
| IHTLS [44] | 4.72 | 8.6 | 4.34 | 1.22 | 12.07 | 2.38 | 5.82 | 1.94 | 1.4 |
| TBD [45] | 5.92 | 10.77 | 5 | 1.99 | 12.75 | 6.55 | 5.9 | 2.62 | 1.79 |
| GOG [24] | 6.16 | 11.03 | 5.3 | 2.14 | 17.05 | 1.8 | 5.67 | 3.7 | 2.55 |
| CMOT [46] | 14.22 | 22.11 | 14.58 | 5.98 | 27.72 | 17.95 | 7.79 | 9.95 | 7.71 |
| Flow-Tracker | **30.87** | **41.84** | **31** | **19.77** | **48.44** | **26.19** | **29.5** | **18.65** | **31.56** |

Table 6. The experimental results on VisDrone2019-MOT test set.

The overall AP increases by 6.5 points and the detection accuracy has a significant improvement. From the comparison in Table 4, we can also get the same conclusion. The number of false alarms drops from 19979 to 10179 and there is also a certain reduction in the number of missing detections. The improvement of the detection results is also beneficial for obtaining better matching results, so the number of ID switches is also greatly reduced. Because the MOTA is closely related to false positives, missing detections and ID switches, so we see that the MOTA has risen from 12.6 to 26.7 in Table 4. The last row of Table 3 and Table 4 shows that we further improve the object detector by using Soft-NMS, deformable convolution and other tricks and it forms our Flow-Tracker eventually. The AP and MOTA have reached the highest level of 21.7 and 32.1 respectively, and almost all the other metrics have certain improvements compared with baseline.

**Speed comparison.** For tracking algorithms, speed is also an important factor we should consider. It is a time consuming process to calculate the optical flow amount of two adjacent frames due to the high resolution of the image, so we employ another method of motion estimation to save time. Specifically, we count the number of unmatched objects in the current frame, and if it exceeds half of the matched objects, we use optical flow estimation to predict the positions of objects at this time. We compare the accuracy and speed of the two methods in Table 5. We can see that the method performing motion estimation per frame has a higher accuracy (AP and MOTA), but its speed is only 5 FPS which cannot achieve real time. Conversely, the other way is much faster, but at the expense of a little accuracy. We can therefore achieve a trade-off between accuracy and speed.

### 4.3. Results on test set

We also report the performance of our method on the VisDrone2019-MOT test set, which is shown in Table 6. We use the improved Cascade R-CNN mentioned above as the detector. The main evaluation metric on test set is the mean average precision (mAP) across object classes over different thresholds. Our Flow-Tracker achieves an AP of 30.87, which far exceeds all the baseline methods and the running speed can reach 5 FPS. What's more, our method achieves the highest accuracy in all categories, which proves the effectiveness of our method strongly. More experimental results and analysis on test set can refer to VisDrone-VDT2018 [47] and VisDrone-MOT2019 [48].

## 5. Conclusion

In this paper, we propose a new multiple object tracking framework based on IoU Tracker, integrating three our proposed modules. In order to solve the mismatch problem caused by dramatic camera motion, we employ an optical flow network to estimate the global motion between adjacent frames, which can also be considered as a predictor of the object position. We tackle the missing detection problem by introducing an auxiliary tracker, which has a good effect on alleviating the problems of ID switches and fragmentations caused by missing detection. Besides, we construct a cascade matching policy using IoU and appearance feature extracted by a residual network, which improves the matching accuracy significantly. We further compare the effects of several object detection algorithms on the tracking results of MOT. The experimental results on the VisDrone2019-MOT dataset confirm the effectiveness of our method. The proposed tracker can also achieve a trade-off between the accuracy and speed.

# References

[1] W. Luo, J. Xing, A. Milan, X. Zhang, W. Liu, X. Zhao, and T. K. Kim. Multiple object tracking: A literature review. *arXiv preprint arXiv:1409.7618*, 2014.

[2] C. Micheloni, G. L. Foresti, C. Piciarelli, and L. Cinque. An autonomous vehicle for video surveillance of indoor environments. *IEEE Transactions on Vehicular Technology*, 56(2), 487-498, 2007.

[3] E. Bochinski, V. Eiselein, and T. Sikora. High-speed tracking-by-detection without using image information. In *IEEE International Conference on Advanced Video and Signal Based Surveillance (ICAVSS)*, pages 1–6, 2017.

[4] N. Wojke, A. Bewley, and D. Paulus. Simple online and realtime tracking with a deep association metric. In *IEEE International Conference on Image Processing (ICIP)*, pages 3645-3649, 2017.

[5] L. Chen, H. Ai, Z. Zhuang, and C. Shang. Real-time multiple people tracking with deeply learned candidate selection and person re-identification. In *ICME*, Vol. 5, p. 8, 2018.

[6] Q. Chu, W. Ouyang, H. Li, X. Wang, B. Liu, and N. Yu. Online multi-object tracking using CNN-based single object tracker with spatial-temporal attention mechanism. In *Proceedings of the IEEE International Conference on Computer Vision (ICCV)*, pages 4836-4845, 2017.

[7] S. Tang, M. Andriluka, B. Andres, and B. Schiele. Multiple people tracking by lifted multicut and person re-identification. In *Proceedings of the IEEE Conference on Computer Vision and Pattern Recognition (CVPR)*, pages 3539-3548, 2017.

[8] D. Sun, X. Yang, M. Y. Liu, and J. Kautz. PWC-Net: CNNs for optical flow using pyramid, warping, and cost volume. In *Proceedings of the IEEE Conference on Computer Vision and Pattern Recognition (CVPR)*, pages 8934-8943, 2018.

[9] P. Zhu, L. Wen, X. Bian, H. Ling, and Q. Hu. Vision meets drones: A challenge. *arXiv preprint arXiv:1804.07437*, 2018.

[10] A. Robicquet, A. Sadeghian, A. Alahi, and S. Savarese. Learning social etiquette: Human trajectory understanding in crowded scenes. In *European Conference on Computer Vision (ECVV)*, pages 549-565, 2016.

[11] S. Li and D. Y. Yeung. Visual object tracking for unmanned aerial vehicles: A benchmark and new motion models. In *Thirty-First AAAI Conference on Artificial Intelligence*, 2017.

[12] Rudolph Emil Kalman. A new approach to linear filtering and prediction problems. *Journal of basic Engineering*, 82(1), 35–45, 1960.

[13] A. Milan, S. H. Rezatofighi, A. Dick, I. Reid, and K. Schindler. Online multi-target tracking using recurrent neural networks. In *Thirty-First AAAI Conference on Artificial Intelligence*, 2017.

[14] A. Sadeghian, A. Alahi, and S. Savarese. Tracking the untrackable: Learning to track multiple cues with long-term dependencies. In *Proceedings of the IEEE International Conference on Computer Vision (ICCV)*, pages 300-311, 2017.

[15] B. D. Lucas and T. Kanade. An iterative image registration technique with an application to stereo vision. In *Proceedings of the 1981 DARPA Image Understanding Workshop*, pages 121-130, 1981.

[16] P. Fischer, A. Dosovitskiy, E. Ilg, P. Hausser, C. Hazirbas, V. Golkov, P. Smagt, D. Cremers, and T. Brox. Flownet: Learning optical flow with convolutional networks. In *Proceedings of the IEEE International Conference on Computer Vision (ICCV)*, pages 2758-2766, 2015.

[17] E. Ilg, N. Mayer, T. Saikia, M. Keuper, A. Dosovitskiy, and T. Brox. Flownet 2.0: Evolution of optical flow estimation with deep networks. In *Proceedings of the IEEE Conference on Computer Vision and Pattern Recognition (CVPR)*, pages 2462-2470, 2017.

[18] W. Choi and S. Savarese. Multiple target tracking in world coordinate with single, minimally calibrated camera. In *European Conference on Computer Vision (ECCV)*, pages 553-567, 2010.

[19] N. Le, A. Heili, and J. M. Odobez. Long-term time-sensitive costs for crf-based tracking by detection. In *European Conference on Computer Vision (ECCV)*, pages 43-51, 2016.

[20] Z. Wu, A. Thangali, S. Sclaroff, and M. Betke. Coupling detection and data association for multiple object tracking. In *2012 IEEE Conference on Computer Vision and Pattern Recognition (CVPR)*, pages 1948-1955, 2012.

[21] S. Pellegrini, A. Ess, K. Schindler, and L. Van Gool. You'll never walk alone: Modeling social behavior for multi-target tracking. In *2009 IEEE 12th International Conference on Computer Vision (ICCV)*, pages 261-268, 2009.

[22] L. Zhao, X. Li, Y. Zhuang, and J. Wang. Deeply-learned part-aligned representations for person re-identification. In *Proceedings of the IEEE International Conference on Computer Vision (ICCV)*, pages 3219-3228, 2017.

[23] L. Leal-Taixé, C. Canton-Ferrer, and K. Schindler. Learning by tracking: Siamese CNN for robust target association. In *Proceedings of the IEEE Conference on Computer Vision and Pattern Recognition (CVPR) Workshops*, pages 33-40, 2016.

[24] H. Pirsiavash, D. Ramanan, and C. C. Fowlkes. Globally-optimal greedy algorithms for tracking a variable number of objects. In *Proceedings of the IEEE Conference on Computer Vision and Pattern Recognition (CVPR)*, pages 1201-1208, 2011.

[25] S. Tang, B. Andres, M. Andriluka, and B. Schiele. Subgraph decomposition for multi-target tracking. In *Proceedings of the IEEE Conference on Computer Vision and Pattern Recognition (CVPR)*, pages 5033-5041, 2015.

[26] S. Tang, B. Andres, M. Andriluka, and B. Schiele. Multi-person tracking by multicut and deep matching. In *European Conference on Computer Vision (ECCV)*, pages 100-111, 2016.

[27] J. Munkres. Algorithms for the assignment and transportation problems. *Journal of the Society for Industrial and Applied Mathematics*, 5(1), 32-38, 1957.

[28] Y. Xu, Y. Ban, X. Alameda-Pineda, and R. Horaud. DeepMOT: A Differentiable Framework for Training Multiple Object Trackers. *arXiv preprint arXiv:1906.06618*, 2019.

[29] P. Felzenszwalb, R. Girshick, D. McAllester, and D. Ramanan. Object detection with discriminatively trained part-based models. *IEEE Transactions on Pattern Analysis and Machine Intelligence*, 32(9), 1627-1645, 2009.

[30] S. Ren, K. He, R. Girshick, and J. Sun. Faster r-cnn: Towards real-time object detection with region proposal networks.

In *Advances in Neural Information Processing Systems (NIPS)*, pages 91-99, 2015.
[31] J. Redmon and A. Farhadi. Yolov3: An incremental improvement. *arXiv preprint arXiv:1804.02767*, 2018.
[32] T. Y. Lin, P. Dollár, R. Girshick, K. He, B. Hariharan, and S. Belongie. Feature pyramid networks for object detection. In *Proceedings of the IEEE Conference on Computer Vision and Pattern Recognition (CVPR)*, pages 2117-2125, 2017.
[33] T. Y. Lin, P. Goyal, R. Girshick, K. He, and P. Dollár. Focal loss for dense object detection. In *Proceedings of the IEEE International Conference on Computer Vision (ICCV)*, pages 2980-2988, 2017.
[34] S. Zhang, L. Wen, X. Bian, Z. Lei, and S. Z. Li. Single-shot refinement neural network for object detection. In *Proceedings of the IEEE Conference on Computer Vision and Pattern Recognition (CVPR)*, pages 4203-4212, 2018.
[35] Z. Cai and N. Vasconcelos. Cascade r-cnn: Delving into high quality object detection. In *Proceedings of the IEEE Conference on Computer Vision and Pattern Recognition (CVPR)*, pages 6154-6162, 2018.
[36] W. Ouyang and X. Wang. Joint deep learning for pedestrian detection. In *Proceedings of the IEEE International Conference on Computer Vision (ICCV)*, pages 2056-2063, 2013.
[37] X. Wang, T. Xiao, Y. Jiang, S. Shao, J. Sun, and C. Shen. Repulsion loss: Detecting pedestrians in a crowd. In *Proceedings of the IEEE Conference on Computer Vision and Pattern Recognition (CVPR)*, pages 7774-7783, 2018.
[38] N. Mayer, E. Ilg, P. Hausser, P. Fischer, D. Cremers, A. Dosovitskiy, and T. Brox. A large dataset to train convolutional networks for disparity, optical flow, and scene flow estimation. In *Proceedings of the IEEE Conference on Computer Vision and Pattern Recognition (CVPR)*, pages 4040-4048, 2016.
[39] K. He, X. Zhang, S. Ren, and J. Sun. Deep residual learning for image recognition. In *Proceedings of the IEEE Conference on Computer Vision and Pattern Recognition (CVPR)*, pages 770-778, 2016.
[40] L. Zheng, L. Shen, L. Tian, S. Wang, J. Wang, and Q. Tian. Scalable person re-identification: A benchmark. In *Proceedings of the IEEE International Conference on Computer Vision (ICCV)*, pages 1116-1124, 2015.
[41] L. Zheng, Z. Bie, Y. Sun, J. Wang, C. Su, S. Wang, and Q. Tian. Mars: A video benchmark for large-scale person re-identification. In *European Conference on Computer Vision (ECCV)*, pages 868-884, 2016.
[42] A. Milan, S. Roth, and K. Schindler. Continuous energy minimization for multitarget tracking. *IEEE Transactions on Pattern Analysis and Machine Intelligence*, *36*(1), 58-72, 2013.
[43] L. Wen, W. Li, J. Yan, Z. Lei, D. Yi, and S. Z. Li. Multiple target tracking based on undirected hierarchical relation hypergraph. In *Proceedings of the IEEE Conference on Computer Vision and Pattern Recognition (CVPR)*, pages 1282-1289, 2014.
[44] C. Dicle, O. I. Camps, and M. Sznaier. The way they move: Tracking multiple targets with similar appearance. In *Proceedings of the IEEE International Conference on Computer Vision (ICCV)*, pages 2304-2311, 2013.
[45] A. Geiger, M. Lauer, C. Wojek, C. Stiller, and R. Urtasun. 3d traffic scene understanding from movable platforms. *IEEE Transactions on Pattern Analysis and Machine Intelligence*, *36*(5), 1012-1025, 2013.
[46] S. H. Bae and K. J. Yoon. Robust online multi-object tracking based on tracklet confidence and online discriminative appearance learning. In *Proceedings of the IEEE Conference on Computer Vision and Pattern Recognition (CVPR)*, pages 1218-1225, 2014.
[47] P. Zhu, L. Wen, D. Du, X. Bian, H. Ling, Q. Hu, ... and X. Liu. VisDrone-VDT2018: The Vision Meets Drone Video Detection and Tracking Challenge Results. In *Proceedings of the European Conference on Computer Vision (ECCV)*, pages 496-518, 2018.
[48] L. Wen, P. Zhu, D. Du, X. Bian, H. Ling, Q. Hu, and *et al*. VisDrone-MOT2019: The Vision Meets Drone Multiple Object Tracking Challenge Results. 2019.